\documentclass[11pt]{article}
\usepackage{coling2020}
\usepackage{times}
\usepackage{url}
\usepackage{latexsym}
\usepackage{graphicx}
\usepackage{amsmath}
\usepackage{multirow}
\usepackage{booktabs}
\usepackage{amsmath} 
\usepackage{amsfonts}
\usepackage{xcolor}
\usepackage{wrapfig}

\colingfinalcopy 

\title{LadaBERT: Lightweight Adaptation of BERT \\through Hybrid Model Compression}

\author{\textbf{Yihuan Mao\textsuperscript{\rm 1,\thanks{~~The work was done when the author visited Microsoft Research Asia.}}, Yujing Wang\textsuperscript{\rm 2,3,\thanks{~~Corresponding Author}}, Chufan Wu\textsuperscript{\rm 1}, Chen Zhang\textsuperscript{\rm 2}, Yang Wang\textsuperscript{\rm 2}}\\
\textbf{~Quanlu Zhang\textsuperscript{\rm 2}, Yaming Yang\textsuperscript{\rm 2}, Yunhai Tong\textsuperscript{\rm 3}, Jing Bai\textsuperscript{\rm 2}}\\
\textsuperscript{\rm 1}Tsinghua University~~
\textsuperscript{\rm 2}Microsoft Research Asia \\
\textsuperscript{\rm 3}Key Laboratory of Machine Perception, MOE, School of EECS, Peking University \\
\tt \small maoyh20@mails.tsinghua.edu.cn, \{yujwang,yhtong\}@pku.edu.cn, chufanwu15@gmail.com \\
\tt \small\{yujwang,zhac,t-yangwa,yayaming,quzha,jbai\}@microsoft.com
}

\begin{document}
\maketitle
\begin{abstract}
BERT is a cutting-edge language representation model pre-trained by a large corpus, which achieves superior performances on various natural language understanding tasks. However, a major blocking issue of applying BERT to online services is that it is memory-intensive and leads to unsatisfactory latency of user requests. Existing solutions leverage knowledge distillation frameworks to learn smaller models that imitate the behaviors of BERT. However, the training procedure of knowledge distillation is expensive itself as it requires sufficient training data to imitate the teacher model. In this paper, we address this issue by proposing a hybrid solution named LadaBERT (Lightweight adaptation of BERT through hybrid model compression), which combines the advantages of different model compression methods, including weight pruning, matrix factorization and knowledge distillation. LadaBERT achieves state-of-the-art accuracy on various public datasets while the training overheads can be reduced by an order of magnitude.
\end{abstract}

\section{Introduction}

The pre-trained language model, BERT~\cite{devlin2018bert} has led to a big breakthrough in various kinds of natural language understanding tasks. Ideally, people can start from a pre-trained BERT checkpoint and fine-tune it on a specific downstream task. However, the original BERT models are memory-exhaustive and latency-prohibitive to be served in embedded devices or CPU-based online environments.
As the memory and latency constraints vary in different scenarios, the pre-trained BERT model should be adaptive to different requirements with accuracy retained to the largest extent. 
Existing BERT-oriented model compression solutions largely depend on knowledge distillation~\cite{hinton2015distilling}, which is inefficient and resource-consuming because a large training corpus is required to learn the behaviors of a teacher. For example, DistilBERT~\cite{sanh2019distilbert} is re-trained on the same corpus as pre-training a vanilla BERT from scratch; and TinyBERT~\cite{jiao2019tinybert} utilizes expensive data augmentation to fit the distillation target. The costs of these model compression methods are as large as pre-training, which are unaffordable for low-resource settings. Therefore, it is straight-forward to ask, \textit{can we design a lightweight method to generate adaptive models with comparable accuracy using significantly less time and resource consumption?} 

In this paper, we propose LadaBERT (Lightweight adaptation of BERT through hybrid model compression) to tackle this problem.
Specifically, LadaBERT is based on an iterative hybrid model compression framework consisting of weighting pruning, matrix factorization and knowledge distillation. Initially, the architecture and weights of student model are inherited from the BERT teacher. In each iteration, the student model is first compressed by a small ratio based on weight pruning and matrix factorization, and is then fine-tuned under the guidance of teacher model through knowledge distillation. Because weight pruning and matrix factorization help to generate better initial and intermediate status for knowledge distillation, both accuracy and efficiency can be greatly improved. 

\begin{wrapfigure}{l}{0.5\textwidth}
    \includegraphics[width=8cm]{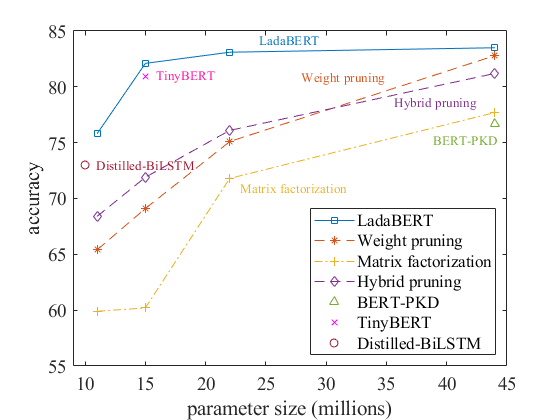}
    \caption{Accuracy comparison on MNLI-m dataset}
    \label{fig:intro}
\end{wrapfigure}

We conduct extensive experiments on five public datasets of natural language understanding. As an example, the performance comparison of LadaBERT and state-of-the-art models on MNLI-m dataset is illustrated in Figure \ref{fig:intro}. 

We can see that LadaBERT outperforms other BERT-oriented model compression baselines at various model compression ratios. Especially, LadaBERT outperforms BERT-PKD significantly under $2.5\times$ compression ratio and outperforms TinyBERT under $7.5\times$ compression ratio while the training speed is accelerated by an order of magnitude. 


The rest of this paper is organized as follows. First, we summarize the related works of model compression and their applications to BERT in Section \ref{sec:related_work}. Then, the methodology of LadaBERT is introduced in Section \ref{sec:ladaBERT}, and experimental results are presented in Section \ref{sec:experiments}. At last, we conclude this work and discuss future works in Section \ref{sec:concolusion}. 


\section{Related Work}
\label{sec:related_work}
Deep Neural Networks (DNNs) have achieved great success in many areas in recent years, but the memory consumption and computational cost expand greatly with the growing complexity of models. Thus, model compression has become an indispensable technique in practice, especially for low-resource scenarios. Here we review the current progresses of model compression techniques briefly, and present their application to pre-trained BERT models.

\subsection{Model compression algorithms}
Existing model compression algorithms can be divided into four categories, namely weight pruning, matrix factorization, weight quantization and knowledge distillation.

Numerous researches have shown that removing a large portion of connections or neurons does not cause significant performance drop in deep neural networks. For example, Han et al.~\shortcite{han2015learning} proposed a method to reduce the storage and computation of neural networks by removing unimportant connections, resulting in sparse networks without affecting the model accuracy. Li et al.~\shortcite{li2016pruning} presented an acceleration method for convolution neural network by pruning whole filters together with their connecting filter maps. This approach does not generate sparse connectivity patterns and brings a much larger acceleration ratio with existing BLAS libraries for dense matrix multiplications. 

Matrix factorization was also widely studied in the deep learning domain, the goal of which is to decompose a matrix into the product of two matrices in lower dimensions.
Sainath et al~\shortcite{sainath2013low} explored a low-rank matrix factorization method of DNN layers for acoustic modeling. Xu et al.~\shortcite{xue2013restructuring,xue2014singular} applied singular value decomposition to deep neural network acoustic models and achieved comparable performances with state-of-the-art models through much fewer parameters. GroupReduce~\cite{chen2018groupreduce} focused on the compression of neural language models and applied low-rank matrix approximation to vocabulary-partition. 
Winata et al.~\shortcite{winata2019effectiveness} carried out experiments for low-rank matrix factorization on different NLP tasks and demonstrated that it was more effective in general than weight pruning. 

Weight quantization is another common technique for compressing deep neural networks, which aims to reduce the number of bits to represent every weight in the model. 
With weight quantization, the weights can be reduced to at most 1-bit binary value from 32-bits floating point numbers. Zhou et al.~\shortcite{zhou2016dorefa} showed that quantizing weights to 8-bits does not hurt the performance; Binarized Neural Networks~\cite{hubara2016binarized} contained binary weights and activations of only one bit; and Incremental Network Quantization~\cite{zhou2017incremental} converted a pre-trained full-precision neural network into low-precision counterpart through three interdependent operations: weight partition, groupwise quantization and re-training. 

Knowledge distillation~\cite{hinton2015distilling} trains a compact and smaller model to approximate the function learned by a large and complex model. A preliminary step of knowledge distillation is to train a deep network (the teacher model) that automatically generates soft labels for training instances. This ``synthetic" label is then used to train a smaller network (the student model), which assimilates the function learned by the teacher model. Chen et al.~\shortcite{chen2017learning} successfully applied knowledge distillation to object detection tasks by introducing several modifications, including a weighted cross-entropy loss, a teacher bounded loss, and adaptation layers to model intermediate teacher distributions. Li et al.~\shortcite{li2017learning} developed a framework to learn from noisy labels, where the knowledge learned from a clean dataset and semantic knowledge graph were leveraged to correct the wrong labels. 

To improve the performance of model compression, there are also numerous attempts to develop hybrid model compression methods that combine more than one category of algorithms. Han et al.~\shortcite{han2016deep} combined quantization, hamming coding and weight pruning to conduct model compression on image classification tasks. Yu et al.~\shortcite{yu2017compressing} proposed a unified framework for low-rank and sparse decomposition of weight matrices with feature map reconstructions. Polino et al.~\shortcite{polino2018model} advocated a combination of distillation and quantization techniques and proposed two hybrid models, \textit{i.e.}, quantified distillation and differentiable quantization. Li et al.,~\shortcite{li2018compression} compressed DNN-based acoustic models through knowledge distillation and pruning. 

\subsection{BERT model compression}
\label{nlp_applications}
In the natural language processing community, there is a growing interest recently to study BERT-oriented model compression for shipping its performance gain into latency-critical or low-resource scenarios. Most existing works focus on knowledge distillation. For instance, BERT-PKD~\cite{sun2019patient} is a patient knowledge distillation approach that compresses the original BERT model into a lightweight shallow network. Different from traditional knowledge distillation methods, BERT-PKD enables an exploitation of rich information in the teacher's hidden layers by utilizing a layer-wise distillation constraint. DistillBERT~\cite{sanh2019distilbert} pre-trains a smaller general-purpose language model on the same corpus as vanilla BERT. Distilled BiLSTM~\cite{tang2019distilling} adopts a single-layer BiLSTM as the student model and achieves comparable results with ELMo~\cite{peters2018deep} through much fewer parameters and less inference time. TinyBERT~\cite{jiao2019tinybert} exploits a novel attention-based distillation schema that encourages the linguistic knowledge in teacher to be well transferred into the student model. It adopts a two-stage learning framework, including general distillation (pre-training from scratch via distillation loss) and task-specific distillation with data augmentation. Both procedures require huge resources and long training times (from several days to weeks), which is cumbersome for industrial applications. Therefore, we are aiming to explore more lightweight solutions in this paper.

\section{Lightweight Adaptation of BERT}
\label{sec:ladaBERT}

\begin{figure*}
\centering
\includegraphics[width=14cm]{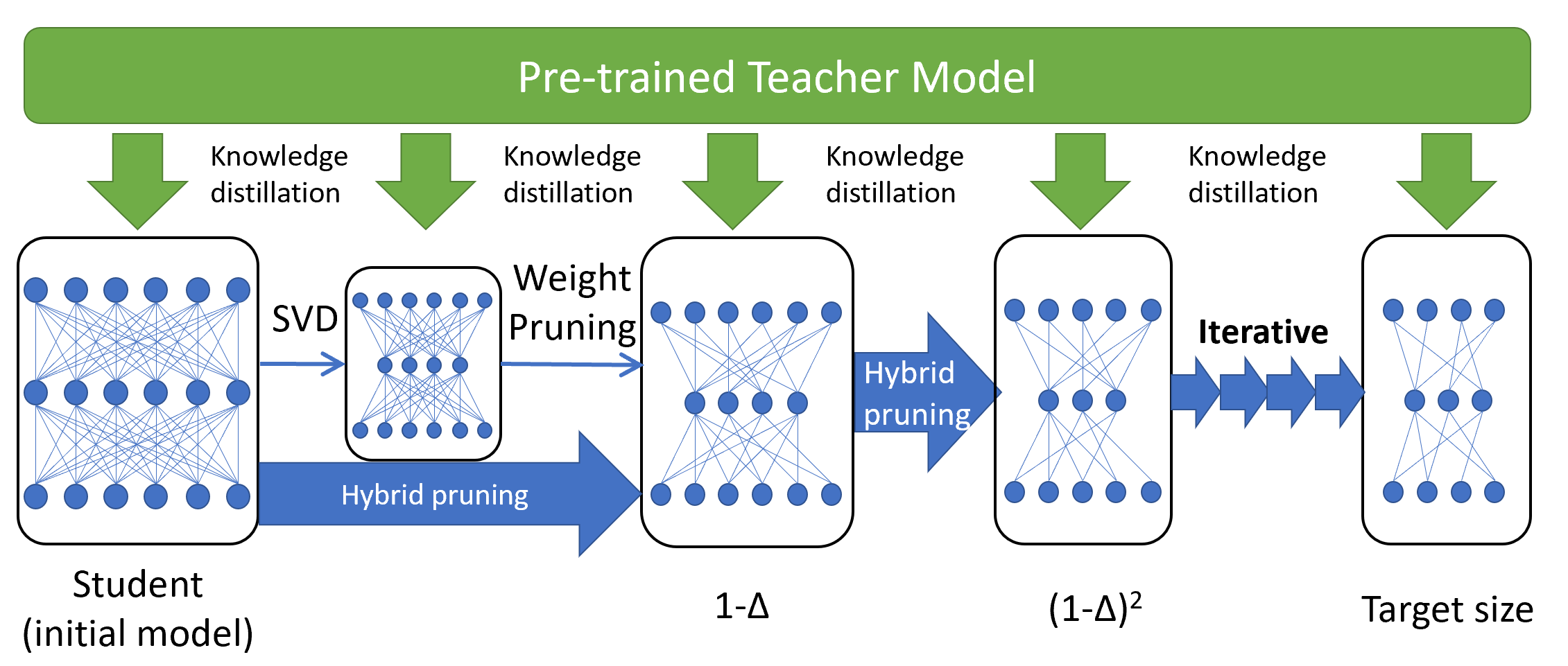}
\caption{\label{fig:overview} Overview of LadaBERT framework}
\end{figure*}

\subsection{Overview}
The overall pipeline of LadaBERT (Lightweight Adaptation of BERT) is illustrated in Figure \ref{fig:overview}. As shown in the figure, the pre-trained BERT model (e.g., BERT-Base) is served as the teacher as well as the initial status of the student model. Then, the student model is compressed towards smaller parameter size iteratively through a hybrid model compression approach until the target size is reached. 
Concretely, in each iteration, the parameter size of student model is first reduced by $1-\Delta$ based on weight pruning and matrix factorization, and then the parameters are fine-tuned by the loss function of knowledge distillation.
The motivation behind is that matrix factorization and weight pruning are complementary to each other. Matrix factorization calculates the optimal approximation under a certain rank, while weight pruning introduces additional sparsity to the decomposed matrices. Moreover, both weight pruning and matrix factorization generate better initial and intermediate status of the student model, which improve the efficiency and effectiveness of knowledge distillation.
In the following subsections, we will introduce the algorithms in detail.

\subsection{Matrix factorization}
\label{svd_pruning}
We use Singular Value Decomposition (SVD) for matrix factorization. All parameter matrices, including the embedding ones, are compressed by SVD. Without loss of generality, we assume a matrix of parameters $\bold{W} \in \mathbb{R}^{m\times n}$, the singular value decomposition of which can be written as:

\begin{equation}
\bold{W}= \bold{U} \Sigma \bold{V}^T
\end{equation}
where $\bold{U} \in \mathbb{R}^{m \times p}$ and $\bold{V} \in \mathbb{R}^{p \times n}$. $\bold{\Sigma} =diag(\sigma_1,\sigma_2,\ldots,\sigma_p)$ is a diagonal matrix composed of singular values and $p$ is the full rank of $W$ satisfying $p \leq min(m, n)$. 


To compress this weight matrix, we select a lower rank $r < p$. The diagonal matrix $\bold{\Sigma}$ is truncated by selecting the top $r$ singular values. i.e., $\bold{\Sigma}_r =diag(\sigma_1, \sigma_2,\ldots,\sigma_r)$, while $\bold{U}$ and $\bold{V}$ are also truncated by selecting the top $r$ columns and rows respectively, resulting in $\bold{U}_r \in\mathbb{R}^{m\times r}$ and $\bold{V}_r \in \mathbb{R}^{r\times n}$. 

Then, low-rank matrix approximation of $\bold{W}$ can be formulated as: 
\begin{equation}
\label{equation_svd}
\hat{\bold{W}}=\bold{U}_r\bold{\Sigma}_r\bold{V}_r^T\ = 
(\bold{U}_r\sqrt{\bold{\Sigma}_r})(\bold{V}_r\sqrt{\bold{\Sigma}_r})^T = \bold{AB}^T
\end{equation}

In this way, the original weight matrix $W$ is decomposed to two smaller matrices, where $\bold{A}=\bold{U}_r\sqrt{\bold{\Sigma}_r} \in \mathbb{R}^{n\times r}$ and $\bold{B}=\bold{V}_r\sqrt{\bold{\Sigma}_r} \in \mathbb{R}^{m\times r}$. These two matrices are initialized by SVD and will be further fine-tuned during training. 


Given a rank $r \leq min(m, n)$, the compression ratio of matrix factorization is defined as: 
\begin{equation}
P_{svd}=\frac{(m+n)r}{mn}
\end{equation}

Therefore, for a target model compression ratio $P_{svd}$, the desired rank $r$ can be calculated by:
\begin{equation}
    r=\frac{mn}{m+n}P_{svd}
\end{equation}

\subsection{Weight pruning}
\label{weight_pruning}
Weight pruning \cite{han2015learning} is an unstructured compression method that induces desirable sparsity for a neural network model. For a neural network $f(\bold{x; \boldsymbol{\theta}})$ with parameters $\boldsymbol{\theta}$, weight pruning finds a binary mask $\bold{M} \in \{0, 1\}^{|\boldsymbol{\theta}|}$ subject to a given sparsity ratio, $P_{weight}$. The neural network after pruning will be $f(\bold{x; M \cdot \boldsymbol{\theta}})$, where the non-zero parameter size is $||\bold{M}||_1 = P_{weight}\cdot|\boldsymbol{\theta}|$, where $|\boldsymbol{\theta}|$ is the number of parameters in $\boldsymbol{\theta}$.
For example, when $P_m = 0.3$, there are 70\% zeros and 30\% ones in the mask $\bold{m}$. In our implementation, we adopt a simple pruning strategy \cite{frankle2018lottery}: the binary mask is generated by setting the smallest weights to zeros.

To combine the benefits of weight pruning and matrix factorization, we leverage a hybrid approach that applies weight pruning on the basis of decomposed matrices generated by SVD. Following Equation (\ref{equation_svd}), SVD-based matrix factorization for any weight matrix $\bold{W}$ can be written as:
$\bold{W}_{svd}=\bold{A}_{m\times r}\bold{B}_{n\times r}^T$.
Then, weight pruning is applied on the decomposed matrices $\bold{A} \in \mathbb{R}^{m \times r}$ and $\bold{B} \in \mathbb{R}^{n \times r}$ separately. The weight matrix after hybrid compression is formulated as:
\begin{equation}
\bold{W}_{hybrid}=(\bold{M_A} \cdot \bold{A})(\bold{M_B}\cdot\bold{B})^T
\end{equation}
where $\bold{M_A}$ and $\bold{M_B}$ are binary masks derived by the weight pruning algorithm with compression ratio $P_{weight}$. 
The compression ratio after hybrid compression can be calculated by:
\begin{equation}
P_{hybrid}=P_{svd}\cdot P_{weight}=\frac{(m+n)r}{mn}P_{weight}
\end{equation}

In LadaBERT, the hybrid compression produce is applied to each layer of the pre-trained BERT model.
Given an overall model compression target $P$, the following constraint should be satisfied:
\begin{equation}
P\cdot|\boldsymbol{\theta}|=P_{embd}\cdot|\boldsymbol{\theta}_{embd}|+P_{hybrid}|\boldsymbol{\theta}_{encd}|+|\boldsymbol{\theta}_{cls}|
\end{equation}
where $|\boldsymbol{\theta}|$ is the total number of model parameters and $P$ is the target compression ratio; $|\boldsymbol{\theta}_{embd}|$ denotes the parameter number of embedding layer, which has a relative compression ratio of $P_{embd}$, and $|\boldsymbol{\theta}_{encd}|$ denotes the number of parameters of all layers in BERT encoder, which have a compression ratio of $P_{hybrid}$. The classification layers (MLP layers with Softmax activation) have a relative small number of parameters ($|\boldsymbol{\theta}_{cls}|$), so they are not modified in model compression. In general, we have three flexible hyper-parameters for fine-grained compression: $P_{embed}$, $P_{svd}$ and $P_{weight}$, which can be optimized by random search on the validation data. 

\subsection{Knowledge distillation}

Knowledge distillation (KD) has been widely used to transfer knowledge from a large teacher model to a smaller student model. In other words, the student model mimics the behavior of the teacher model by minimizing the knowledge distillation loss functions. 
Various types of knowledge distillation can be employed at different sub-layers. 
Generally, all types of knowledge distillation can be modeled as minimizing the following loss function:

\begin{equation}
\mathcal{L}_{\mathrm{KD}}=\sum_{\bold{x} \in \mathcal{X}} L\left(f^{(s)}(\bold{x}), f^{(t)}(\bold{x})\right)
\end{equation}
Where  $\mathcal{X}$ denotes the training set and $\bold{x}$ is a sample input in the set. $f^{(s)}(\bold{x})$ and $f^{(t)}(\bold{x})$ represent intermediate outputs or weight matrices for the student model and teacher model respectively. $L(\cdot)$ represents for a loss function which can be carefully designed for different types of knowledge distillation. We partly follow the recent technique proposed by TinyBERT~\cite{jiao2019tinybert}, which applies knowledge distillation constraints upon embedding, self-attention, hidden representation and prediction levels. Concretely, there are four types of knowledge distillation constraints as follows:
\begin{itemize}

    \item \textbf{Embedding-layer distillation} is performed upon the embedding layer. $f(\bold{x}) \in \mathbb{R}^{n \times d}$ represents for the word embedding output for input $x$, where $n$ is the input word length and $d$ is the dimension of word embedding. Mean Squared Error (MSE) is adopted as the loss function $L(\cdot)$. 
\item \textbf{Attention-layer distillation} is performed upon the self-attention sub-layer. $f(\bold{x}) = \{a_{ij}\}\in \mathbb{R}^{n \times n}$ represents the attention output for each self-attention sub-layer, and $L(\cdot)$ denotes MSE loss function. 

\item \textbf{Hidden-layer distillation} is performed at each fully-connected sub-layer in the Transformer architectures. $f(\bold{x})$ denotes the output representation of the corresponding sub-layer, and $L(\cdot)$ also adopts MSE loss function. 

\item \textbf{Prediction-layer distillation} makes the student model to learns the predictions from a teacher model directly. It is identical to a vanilla form of knowledge distillation~\cite{hinton2015distilling}. It takes soft cross-entropy loss function, which can be formulated as:
    \begin{equation}
    \mathcal{L}_{\mathrm{pred}}=-\sigma(f^t(\bold{x})) \cdot \log{(\sigma(f^s(\bold{x})/t))}
    \end{equation}
    where $\sigma(\cdot)$ denotes Softmax function, $f^t(\bold{x})$ and $f^s(\bold{x})$ are the predictive logits of teacher and student models respectively. $t$ is a temperature value, which generally works well at $t=1$~\cite{jiao2019tinybert}.
\end{itemize}

\section{Experiments}
\label{sec:experiments}

\subsection{Datasets \& Baselines}
We compare LadaBERT with state-of-the-art model compression approaches on five public datasets of different tasks of natural language understanding, including sentiment classification (SST-2), natural language inference (MNLI-m, MNLI-mm, QNLI) and pairwise semantic equivalence (QQP). The statistics of these datasets are described in Table \ref{tab:Statistics}. 


\begin{table}[ht]
    \renewcommand\arraystretch{0.9}
    \centering
    \begin{tabular}{ccccc}
    \hline
     Task & \#Train & \#Dev. & \#Test & \#Class \\ \hline
     SST-2 & 67,350 & 873 & 1,822 & 2    \\
     QQP & 363,871 & 40,432 & 390,965 & 2  \\
     MNLI-m & 392,703 & 9,816 & 9,797 & 3  \\
     MNLI-mm & 392,703 & 9,833 & 9,848 & 3  \\
     QNLI & 104,744 & 5,464 & 5,464 & 2  \\
     \hline
    \end{tabular}
    \caption{Dataset Statistics}
    \label{tab:Statistics}
\end{table}


The baseline approaches are summarized below.
\begin{itemize}
    \item \textbf{Weight pruning} and \textbf{Matrix factorization} are two simple baselines described in Section \ref{weight_pruning}. We evaluate both pruning methods in an iterative manner until the target compression ratio is reached.
    
    \item \textbf{Hybrid pruning} is a combination of matrix factorization and weight pruning, which conducts iterative weight pruning on the basis of SVD-based matrix factorization. It is performed iteratively until the desired compression ratio is achieved. 
    
    \item \textbf{BERT-FT}, \textbf{BERT-KD} and \textbf{BERT-PKD} are reported in \cite{sun2019patient}, where BERT-FT directly fine-tunes the model via supervision labels, BERT-KD is the vanilla knowledge distillation algorithm~\cite{hinton2015distilling}, and BERT-PKD stands for Patient Knowledge Distillation proposed in \cite{sun2019patient}. The student model is composed of 3 Transformer layers, resulting in a $2.5\times$ compression ratio. Each layer has the same hidden size as the pre-trained teacher, so the initial parameters of student model can be inherited from the corresponding teacher. 
    
    \item \textbf{TinyBERT} \cite{jiao2019tinybert} instantiates a tiny student model, which has totally 14.5M parameters ($7.5\times$ compression ratio) composed of 4 layers, 312 hidden units, 1200 intermediate size and 12 heads. For a fair comparison, we reproduce the TinyBERT pipeline\footnote{https://github.com/huawei-noah/Pretrained-Language-Model/tree/master/TinyBERT} without general distillation and data augmentation, which is time-exhaustive and resource-consuming.
    
    \item \textbf{BERT-Small} has the same model architecture as TinyBERT, but is directly pre-trained by the official BERT pipeline. The performance values are copied from \cite{jiao2019tinybert} for reference.
    
    \item \textbf{Distilled-BiLSTM}~\cite{tang2019distilling} leverages a single-layer bidirectional-LSTM as the student model, where the hidden units and intermediate size are set to be 300 and 400 respectively, resulting in a $10.8 \times$ compression ratio. This model requires an expensive training process similar to vanilla BERT.
    \end{itemize}


\subsection{Setup}
We leverage the pre-trained checkpoint of \textit{base-bert-uncased}\footnote{https://storage.googleapis.com/bert\_models/2018\_10\_18/uncased\_L\-12\_H\-768\_A\-12.zip} as the initial model for compression, which contains 12 layers, 12 heads, 110M parameters, and 768 hidden units per layer. Hyper-parameter selection is conducted on the validation data for each dataset. After training, the prediction results are submitted to the GLUE-benchmark evaluation platform\footnote{https://gluebenchmark.com/} to get the evaluation performance on test data.

For a comprehensive evaluation, we experiment with four settings of LadaBERT, namely LadaBERT-1, -2, -3 and -4, which reduce the model parameters of  BERT-Base by 2.5, 5.0, 7.5 and 10.0 times respectively. 
In our experiment, we set the batch size as 32 and learning rate as 2e-5. The optimizer is BertAdam with the default setting~\cite{devlin2018bert}. Fine-grained compression ratios are optimized by random search on SST dataset and transferred to other datasets (shown in Table \ref{table:params}). Following \cite{jiao2019tinybert}, the temperature value in distillation loss function is set as 1 in all experiments without tuning. 

\begin{table*}[!h]
 \centering
 \caption{Fine-grained compression ratios}
 \scalebox{0.9}{
  \begin{tabular}{c|c|c|c|c}
  \toprule
    \textbf{Model} & \textbf{Overall} & \textbf{Embedding layer} & \textbf{Matrix factorization} & \textbf{Weight pruning}\\
    \midrule
    LadaBERT-1 & $\times 2.5$ & $\times 1.43$ & $\times 2.0$ & $\times 1.56$\\
    LadaBERT-2 & $\times 5.0$ & $\times 2.05$ & $\times 2.0$ & $\times 3.41$\\
    LadaBERT-3 & $\times 7.5$ & $\times 5.0$ & $\times 2.0$ & $\times 4.33$\\
    LadaBERT-4 & $\times 10.0$ & $\times 5.0$ & $\times 2.5$ & $\times 5.45$\\
    \bottomrule
  \end{tabular}
 }
\label{table:params}
\end{table*}

\subsection{Performance Comparison}

\begin{table*}[!h]
 \centering
 \caption{Performance comparison on various model sizes}
 \scalebox{0.9}{
  \begin{tabular}{l|ccccc|c|c}
  \toprule
    \textbf{Algorithm} & \textbf{MNLI-m} & \textbf{MNLI-mm} & \textbf{SST-2} & \textbf{QQP} & \textbf{QNLI} & \textbf{\#Params} & \textbf{Ratio} \\ \midrule
    \textbf{BERT-Base} & \textbf{84.6} & \textbf{83.4} & \textbf{93.5} & \textbf{71.2/-} & \textbf{90.5} & 110M & $\times$1.0 \\ \midrule
    \textbf{LadaBERT-1} & \textbf{83.5} & \textbf{82.5} & \textbf{92.8} & \textbf{70.7/88.9} & \textbf{89.6} & 44M & $\times$2.5 \\ 
    BERT-FT & 74.8 & 74.3 & 86.4& 65.8/86.9& 84.3& 44M & $\times$2.5 \\
    BERT-KD & 75.4 & 74.8 & 86.9 & 67.3/87.6 & 84.0 & 44M &  $\times$2.5 \\ 
    BERT-PKD & 76.7 & 76.3 & 87.5 & 68.1/87.8 & 84.7 & 44M & $\times$2.5 \\
    Weight pruning & 82.8 & 81.6 & 92.3 & 70.1/88.5 & 88.9 & 44M & $\times$2.5 \\
    matrix factorization & 77.7 & 77.4 & 87.6 & 65.7/87.2 & 84.3 & 44M & $\times$2.5 \\
    Hybrid pruning & 81.2 & 80.0 & 90.0 & 68.0/87.5 & 83.3 & 44M & $\times$2.5 \\
    \midrule
    \textbf{LadaBERT-2} & \textbf{83.1} & \textbf{82.2} & \textbf{91.8} & \textbf{69.9/87.9} & \textbf{88.2} & 22M & $\times$5.0 \\
    Weight pruning & 75.9 & 75.6 & 84.8 & 60.3/83.5 & 81.7 & 22M & $\times$5.0 \\
    matrix factorization & 71.8 & 71.8 & 82.8 & 60.3/83.5 & 75.4 & 22M & $\times$5.0 \\
    Hybrid pruning & 76.1 & 75.3 & 85.4 & 64.9/85.8 & 80.6 & 22M & $\times$5.0 \\
    \midrule
    \textbf{LadaBERT-3} & \textbf{82.1} & \textbf{81.8} & \textbf{89.9} & \textbf{69.4/87.8} & 84.5 & 15M & $\times$7.5 \\
    TinyBERT & 80.9 & 79.5 & 89.5 & 65.4/87.5 & 77.9 & 15M & $\times$7.5 \\
    BERT-Small & 75.4 & 74.9 & 87.6 & 66.5/- & \textbf{84.8}& 15M & $\times$7.5 \\
    Weight pruning & 69.1 & 68.8 & 81.8 & 59.7/82.9 & 76.4 & 15M & $\times$7.5 \\
    matrix factorization & 60.2 & 60.0 & 81.3 & 58.5/82.0 & 62.2 & 15M & $\times$7.5 \\
    Hybrid pruning & 71.9 & 71.0 & 83.5 & 62.3/84.7 & 73.8 & 15M & $\times$7.5 \\
    \midrule
    \textbf{LadaBERT-4} & \textbf{75.8} & \textbf{76.1}  & 84.0  & 67.4/86.6 & \textbf{75.1} & 11M & $\times$10.0 \\
    Distilled-BiLSTM & 73.0& 72.6 & \textbf{90.7} & \textbf{68.2/88.1} & - & 10M & $\times$10.8 \\
    Weight pruning & 64.9  & 65.1 & 80.4 & 56.9/80.5 & 62.7 & 11M & $\times$10.0 \\
    matrix factorization & 59.9 & 59.6 & 79.2 & 57.8/81.9 & 62.2 & 11M & $\times$10.0 \\
    Hybrid pruning & 68.4 & 67.9 & 81.5 & 58.6/83.5 & 63.2 & 11M & $\times$10.0 \\
    \bottomrule
  \end{tabular}
 }
\label{table:overall}
\end{table*}

The evaluation results of LadaBERT and state-of-the-art approaches are listed in Table \ref{table:overall}, where the models are ranked by parameter sizes for feasible comparison. As shown in the table, LadaBERT consistently outperforms the strongest baselines under similar model sizes. In addition, the performance of LadaBERT demonstrates the superiority of a combination of SVD-based matrix factorization, weight pruning and knowledge distillation. 

With model size of $2.5\times$ reduction, LadaBERT-1 performs significantly better than BERT-PKD, boosting the performance by relative 8.9, 8.1, 6.1, 3.8 and 5.8 percentages on MNLI-m, MNLI-mm, SST-2, QQP and QNLI datasets respectively. 
Recall that BERT-PKD initializes the student model by selecting 3 of 12 layers in the pre-trained BERT-Base model. It turns out that the discarded layers have a huge impact on the model performance, which is hard to be recovered by knowledge distillation. On the other hand, LadaBERT generates the student model by iterative pruning on the pre-trained teacher. In this way, the original knowledge in the teacher model can be preserved to the largest extent, and the benefit is complementary to knowledge distillation.

LadaBERT-3 has a comparable size as TinyBERT with a $7.5 \times$ compression ratio. As shown in the results, TinyBERT does not work well without expensive data augmentation and general distillation, hindering its application to low-resource settings.
The reason is that the student model of TinyBERT is distilled from scratch, so it requires much more data to mimic the teacher's behaviors. Instead, LadaBERT has better initial and intermediate status calculated by hybrid model compression, which is much more light-weighted and achieves competitive performances with much faster learning speed (learning curve comparison is shown in Section \ref{sec:learning_curve}). 
Moreover, LadaBERT-3 outperforms BERT-Small on most of the datasets, which is pre-trained from scratch by the official BERT pipeline. This means that LadaBERT can quickly adapt to smaller model sizes and achieve competitive performance without expansive re-training on a large corpus. 


Moreover, Distilled-BiLSTM performs well on SST-2 dataset with more than $10 \times$ compression ratio, owing to good generalization ability of LSTM model on small datasets. Nevertheless, the performance of LadaBERT-4 is competitive on larger datasets such as MNLI and QQP. This is impressive as LadaBERT is much more efficient without exhaustive re-training on a large corpus. In addition, the inference speed of BiLSTM is slower than transformer-based models with similar parameter sizes. 


\subsection{Learning curve comparison}
\label{sec:learning_curve}
\begin{figure}[t]
\begin{minipage}{0.45\textwidth}
\centering
\includegraphics[width=\textwidth]{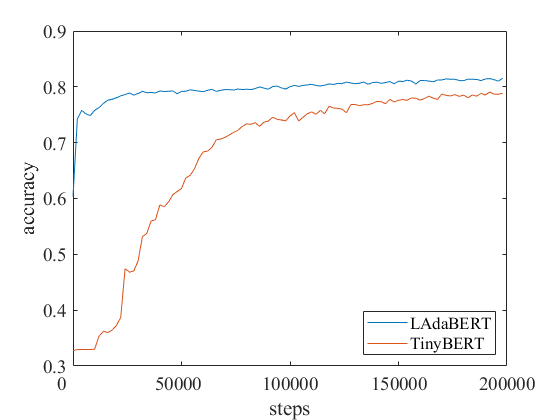}
\caption{Learning curve on MNLI-m dataset.}
\label{fig:learning_curve_1}
\end{minipage}
\begin{minipage}{0.45\textwidth}
\centering
\includegraphics[width=\textwidth]{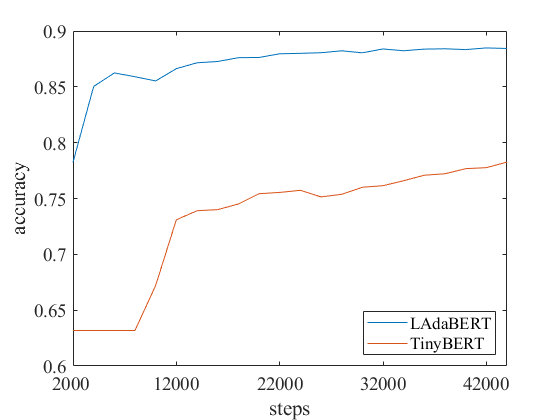}
\caption{Learning curve on QQP dataset.}
\label{fig:learning_curve_2}
\end{minipage}
\end{figure}

To further demonstrate the efficiency of LadaBERT, we visualize the learning curves on MNLI-m and QQP datasets in Figure \ref{fig:learning_curve_1} and \ref{fig:learning_curve_2}, where LadaBERT-3 is compared to the strongest baseline, TinyBERT, under $7.5 \times$ compression ratio. 
As shown in the figures, LadaBERT-3 achieves good performances much faster and results in a better convergence point. After training $2 \times 10^4$ steps (batches) on MNLI-m dataset, the performance of LadaBERT-3 is already comparable to TinyBERT after convergence (approximately $2 \times 10^5$ steps), achieving nearly $10$ times acceleration. And on QQP dataset, both performance improvement and training speed acceleration are very significant. This clearly shows the superiority of combining matrix factorization, weight pruning and knowledge distillation in a collaborative manner. On the other hand, TinyBERT is based on pure knowledge distillation, so the learning speed is much slower. 


\subsection{Effect of low-rank + sparsity}
In this paper, we demonstrate that a combination of matrix factorization and weight pruning is better than single solutions for BERT-oriented model compression. 
Similar phenomena has been reported in computer vision, showing that low-rank and sparsity are complementary to each other~\cite{yu2017compressing}. Here we provide another explanation to support our observation.

\begin{wrapfigure}{l}{0.5\textwidth}
    \includegraphics[width=7cm]{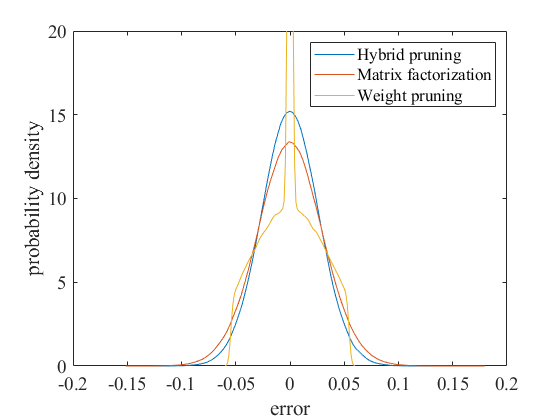}
    \caption{Distribution of pruning errors}
    \label{fig:hybrid_anal}
\end{wrapfigure}

In Figure \ref{fig:hybrid_anal}, we visualize the distribution of element biases for a weight matrix in the neural network after pruning to 20\% of its original parameter size. 
For illustration, we consider the matrix initialized by real pretrained BERT weights, and the pruning process is done at once.  
We define the biases to be calculated by $\mathop{\mathbf{Bias}}_{ij}=\hat{\bold{M}}_{ij}-\bold{M}_{ij}$, where $\hat{\bold{M}}$ denotes the weight matrix after pruning.


The yellow line in Figure \ref{fig:hybrid_anal} shows the distribution of biases generated by pure weight pruning, which has a sudden drop at the pruning threshold. The orange line represents for pure SVD pruning, which turns out to be smoother and is aligned with Gaussian distribution. The blue line shows the result of hybrid pruning, which conducts weight pruning on the decomposed matrices. First, we apply SVD-based matrix factorization to reduce 60\% of total parameters. Then, weight pruning is applied on the decomposed matrices by 50\%, resulting in 20\% parameters while the bias distribution changes slightly. As visualized in Figure \ref{fig:hybrid_anal}, it has smaller mean and deviation of bias distribution than that of pure matrix factorization. In addition, it seems that a smoother weight distribution is more feasible for the fine-tuning procedure. Therefore, it is reasonable that a hybrid model compression approach is advantageous than pure weight pruning. 

\section{Conclusion}
\label{sec:concolusion}
Model compression is a common way to deal with latency-critical or memory-intensive scenarios. Existing model compression methods for BERT are expansive as they require re-training on a large corpus to reserve the original performance. In this paper, we propose LadaBERT, a lightweight model compression pipeline that generates an adaptive BERT model efficiently based on a given task and specific constraint. It is based on a hybrid solution, which conducts matrix factorization, weight pruning and knowledge distillation in a collaborative fashion. The experimental results demonstrate that LadaBERT is able to achieve comparable performance with other state-of-the-art solutions using much less training data and computation budget. Therefore, LadaBERT can be easily plugged into various applications to achieve competitive performances with little training overheads. In the future, we would like to apply LadaBERT to large-scale industrial applications, such as search relevance and query recommendation. 



\bibliographystyle{coling}
\bibliography{coling2020}

\end{document}